\documentclass{article}
\usepackage{iclr2026_delta,times}

\usepackage{amsmath,amsfonts,bm}

\def\eqref#1{equation~\ref{#1}}

\def\1{\bm{1}}

\DeclareMathAlphabet{\mathsfit}{\encodingdefault}{\sfdefault}{m}{sl}
\SetMathAlphabet{\mathsfit}{bold}{\encodingdefault}{\sfdefault}{bx}{n}

\DeclareMathOperator*{\argmax}{arg\,max}

\usepackage{hyperref}
\usepackage{url}
\usepackage{graphicx}
\usepackage{booktabs}
\usepackage{algorithm}
\usepackage{algorithmic}
\usepackage{tikz}
\usetikzlibrary{shapes.geometric, arrows.meta, positioning, fit, calc, backgrounds}
\usepackage{listings}
\usepackage[htt]{hyphenat}

\title{SYNTHONY: A Stress-Aware, Intent-Conditioned Agent for Deep Tabular Generative Models Selection}

\author{Hochan Son\thanks{Equal contribution.} \\
Department of Statistics\\
University of California, Los Angeles\\
\texttt{hochanson@g.ucla.edu} \\
\And
Xiaofeng Lin\footnotemark[1] \\
Department of Statistics\\
University of California, Los Angeles\\
\texttt{bernardo1998@ucla.edu} \\
\And
Jason Ni \\
Department of Mathematics\\
University of California, Los Angeles\\
\texttt{jasonni19@g.ucla.edu} \\
\And
Guang Cheng \\
Department of Statistics\\
University of California, Los Angeles\\
\texttt{guangcheng@stat.ucla.edu} \\
}

\iclrfinalcopy 

\begin{document}

\maketitle


\begin{abstract}
Deep generative models for tabular data (GANs, diffusion models, and LLM-based generators) exhibit highly non-uniform behavior across datasets; the best-performing synthesizer family depends strongly on distributional stressors such as long-tailed marginals, high-cardinality categoricals, Zipfian imbalance, and small-sample regimes. This brittleness makes practical deployment challenging, especially when users must balance competing objectives of fidelity, privacy, and utility.

We study \emph{intent-conditioned tabular synthesis selection}: given a dataset and a user intent expressed as a preference over evaluation metrics, the goal is to select a synthesizer that minimizes regret relative to an intent-specific oracle. We propose \textbf{stress profiling}, a synthesis-specific meta-feature representation that quantifies dataset difficulty along four interpretable stress dimensions, and integrate it into \textbf{SYNTHONY}, a selection framework that matches stress profiles against a calibrated capability registry of synthesizer families. Across a benchmark of 7 datasets, 10 synthesizers, and 3 intents, we demonstrate that stress-based meta-features are highly predictive of synthesizer performance: a $k$NN selector using these features achieves strong Top-1 selection accuracy, substantially outperforming zero-shot LLM selectors and random baselines. We analyze the gap between meta-feature-based and capability-based selection, identifying the hand-crafted capability registry as the primary bottleneck and motivating learned capability representations as a direction for future work.
\end{abstract}

\section{Introduction}
\label{sec:introduction}

The central challenge in tabular data synthesis is no longer \textit{access} to generative models, but the \textit{selection} of them. GANs, diffusion models, and LLM-based generators now offer diverse architectures for producing synthetic tabular data \citep{xu2019ctgan,kotelnikov2023tabddpm,borisov2022great}, addressing data scarcity, privacy constraints, and class imbalance \citep{bauer2024sdg_survey,chawla2002smote}. Yet unlike supervised tabular learning, where AutoML frameworks automate the search from preprocessing to hyperparameter tuning \citep{feurer2015autosklearn,erickson2020autogluon}, the synthesis landscape lacks a comparable ``Auto-Synthesizer'' paradigm. Practitioners must manually select between architectures ranging from copulas to diffusion models, a choice that is particularly risky when an untuned complex model (e.g., a GAN) underperforms a simple, well-tuned baseline due to training instability or mode collapse \citep{mescheder2018gan}.

We identify four canonical failure modes that expose the limitations of
uninformed model selection:

\begin{itemize}
  \item \textbf{The Long Tail Problem}: Datasets with severe distributional skew cause mode collapse in basic GANs and VAEs, which optimize for central tendencies rather than tail fidelity.
  
  \item \textbf{The Zipfian Imbalance Problem}: Categorical columns following power-law distributions \citep{newman2005power} cause generative models with continuous latent bottlenecks to silently erase rare-but-critical tail categories---a failure invisible to aggregate fidelity metrics yet catastrophic for minority-sensitive downstream tasks. 

  \item \textbf{The Needle in the Haystack Problem}: High-cardinality categorical variables (hundreds or thousands of unique values) overwhelm embedding-based encoders, leading to mode collapse even when the underlying distribution is not heavily skewed.
  
  \item \textbf{The Small Data Trap}: With limited training samples, expressive deep models memorize rather than generalize, producing synthetic records that leak private information \citep{stadler2022groundhog}.
\end{itemize}
These failure modes are not random -- they are predictable from measurable dataset properties, which motivates a profiling-based approach to selection.

Current open-source libraries, such as SDV \citep{patki2016sdv} and SynthCity \citep{qian2023synthcity}, have made significant strides in democratizing access to individual synthesizers. However, they function as collections of independent building blocks rather than intelligent selection systems. Users must still manually navigate the trade-offs between model expressiveness, computational cost, and data compatibility, a task requiring deep expertise in both generative modeling and dataset characteristics.

To bridge this gap, we study \emph{intent-conditioned synthesis selection} for deep tabular generative modeling. The core hypothesis is that tabular DGMs fail in systematic, predictable ways that depend on dataset stressors (e.g., long tails, high-cardinality categoricals, Zipfian imbalance, and small-sample regimes). We operationalize this hypothesis via \textbf{stress profiling}: a synthesis-specific meta-feature representation that quantifies dataset difficulty along four interpretable dimensions and maps them to known failure modes. We then build \textbf{SYNTHONY}, a selection framework that matches stress profiles against a calibrated capability registry to rank synthesizer families for a given dataset and intent.

Our contributions are:
\begin{itemize}
    \item \textbf{Problem formulation and evaluation.} We formalize \emph{intent-conditioned tabular synthesis selection} and evaluate selectors using Top-$K$ accuracy, Spearman rank correlation, and NDCG against intent-specific oracles.
    \item \textbf{Stress profiling as a predictive principle.} We introduce \emph{stress profiles}, four synthesis-specific, interpretable dataset stress dimensions, and demonstrate empirically that they predict which deep generative model families succeed or fail on tabular data.
    \item \textbf{Capability-based selection framework.} We propose SYNTHONY, which uses stress profiles to score candidate model families via a capability registry grounded in architectural properties and synthesis literature. The registry encodes domain knowledge (e.g., ``diffusion models handle skew better than basic VAEs'') as structured, interpretable scores rather than opaque learned weights, enabling practitioners to inspect and override individual entries. Intent-conditioned scale factors are calibrated from data via Bayesian optimization. We further analyze the gap between meta-feature-based and capability-based selection, identifying directions for learned capability representations.
    \item \textbf{Cross-family empirical analysis.} We benchmark 10 synthesizers across 4 families, 7 datasets, and 3 user intents, revealing that optimal model selection is strongly intent-dependent and that stress-based meta-features capture this dependence more reliably than heuristic rules or zero-shot LLM reasoning.
\end{itemize}

\section{Related Work}
\label{sec:related_work}

\subsection{Generative Models for Tabular Data}
The landscape of tabular synthesis has evolved from statistical methods (Bayesian Networks \citep{zhang2014privbayes}, Gaussian Copulas \citep{patki2016sdv}) through GANs adapted for mixed-type data such as CTGAN \citep{xu2019ctgan} and TableGAN \citep{mahmood2023tablegan}, to diffusion models that now represent the state-of-the-art: TabDDPM \citep{kotelnikov2023tabddpm} introduced cascading Gaussian and multinomial diffusion for mixed types; TabSyn \citep{zhang2024tabsyn} performs diffusion in a VAE-crafted latent space with significant gains in quality and speed. Concurrently, LLMs have been adapted for tabular synthesis (GReaT, REaLTabFormer \citep{borisov2022great,solatorio2023realtabformer}). This architectural diversity creates a selection challenge: without systematic guidance, choosing the optimal generator for a given dataset and constraints remains trial-and-error.

\subsection{Benchmarking and Evaluation}
Existing synthesis libraries focus on \textit{standardization} and \textit{evaluation} rather than automated \textit{selection}. SDV \citep{patki2016sdv} and SynthCity \citep{qian2023synthcity} provide unified APIs and benchmarking but require manual hyperparameter specification. SynMeter \citep{du2024synmeter} and STNG \citep{rashidi2024stng} integrate tuning loops into evaluation, yet still leave model \emph{selection} to the user. Our work introduces an automated selection layer: unlike these tools, \textsc{Synthony} automates the upstream decision of \textit{which} model to use by matching dataset stress profiles against a calibrated capability registry conditioned on user intent.

\subsection{Meta-Learning and Algorithm Selection}
Algorithm selection---the problem of choosing the best solver based on instance features--- was formalized by \citet{rice1976algorithm} and is mature for supervised learning \citep{vanschoren2019meta}, where dataset meta-features enable warm-started search in Auto-sklearn \citep{feurer2015autosklearn} and AutoGluon \citep{erickson2020autogluon} without exhaustive evaluation \citep{rivolli2022meta}. However, this paradigm has not been applied to synthesis. Recent surveys consistently identify synthesizer selection as an open problem: \citet{kindji2025tabular} find rankings shift with compute budget but offer no predictive mechanism; \citet{dsj2025benchmarking} conclude practitioners must manually interpret results; and \citet{davila2024navigating} construct a qualitative guide but stop short of automation. Our work addresses this gap with \textbf{stress profiling}, a synthesis-specific meta-feature framework quantifying dataset difficulty along four dimensions corresponding to known generative failure modes, and \textsc{Synthony}, which uses these profiles for predictive, intent-conditioned selection, complementing existing libraries \citep{patki2016sdv,qian2023synthcity} by automating the question of \textit{which} synthesizer to deploy.

\section{Methodology}
\label{sec:method}

\subsection{Problem Formulation}
We define the tabular synthesis selection problem as a tuple $\mathcal{T} = (\mathcal{D}, \mathcal{I}, \mathcal{G})$, where $\mathcal{D}$ is the private source dataset, $\mathcal{G} = \{g_1, \ldots, g_k\}$ is a set of candidate generative architectures (e.g., GANs, diffusion models, copulas), and $\mathcal{I}$ represents the user's intent. The intent $\mathcal{I}$ specifies a preference over competing evaluation dimensions such as fidelity, privacy, and utility.

Our objective is to identify the optimal generator $g^*$ such that:
\begin{equation}
    g^* = \argmax_{g \in \mathcal{G}} \; \mathcal{F}(\mathbf{m}_g \mid \mathcal{I})
    \label{eq:oracle}
\end{equation}
where $\mathbf{m}_g$ is the evaluation metric vector for generator $g$ on dataset $\mathcal{D}$, and $\mathcal{F}$ is a scalarization function derived from the user's intent $\mathcal{I}$.

Equation~\ref{eq:oracle} defines the \emph{oracle} selection, which requires evaluating every generator. Since this is prohibitively expensive, SYNTHONY approximates it via a surrogate scoring function (Equation~\ref{eq:score}) that operates on pre-computed capability scores rather than evaluation metrics.

\subsection{Stress Profiling: Dataset Difficulty Signatures}
\label{sec:stress-profiling}
Given a dataset $D$ with mixed continuous and categorical features, we compute a stress profile
$\phi(D) \in \mathbb{R}^4$ capturing four synthesis-relevant stressors:
(i)~\textbf{Long-tail skew}: maximum absolute Fisher--Pearson skewness across numeric columns (flagged if any column exceeds $|\text{skew}| > 2.0$);
(ii)~\textbf{Cardinality complexity}: maximum unique-value count across all columns (flagged if any column exceeds 500 unique values);
(iii)~\textbf{Zipfian concentration}: maximum top-20\% frequency ratio across categorical columns, i.e., the fraction of rows occupied by the most frequent 20\% of categories (flagged as high-stress if $> 0.80$);
and (iv)~\textbf{Small-sample regime}: total row count (flagged if $< 500$).
These stressors correspond directly to the failure modes identified in Section~\ref{sec:introduction}: long-tail skew maps to the Long Tail Problem, the Needle in Haystack Problem maps jointly to cardinality complexity and Zipf concentration, and the Small Data Trap maps to the small-sample regime.

Stress profiles serve two roles: (1) they enable interpretable diagnosis of why a synthesizer fails on a given dataset, and
(2) they provide meta-features for predicting promising model families under a fixed evaluation budget.

\paragraph{From stress profiles to requirement vectors.}
The 4-dimensional stress profile $\phi(D) \in \mathbb{R}^4$ is mapped to a 6-dimensional requirement vector $\mathbf{r} \in \{0,\ldots,4\}^6$ used for capability scoring. Each raw stress statistic is quantized into a discrete requirement level via threshold-based binning. For example, long-tail skew maps to \texttt{skew\_handling} as: $r=0$ if $|\text{skew}| < 0.5$ (no stress), $r=1$ if $< 1.0$, $r=2$ if $< 2.0$, $r=3$ if $< 3.0$, and $r=4$ if $\geq 3.0$. Cardinality complexity maps to \texttt{cardinality\_handling} via unique-value-count bins ($<50 \to 0$, $<200 \to 1$, $<500 \to 2$, $<1000 \to 3$, $\geq 1000 \to 4$). Zipf concentration maps to \texttt{zipfian\_handling} via top-20\% frequency ratio bins ($<0.5 \to 0$, $<0.65 \to 1$, $<0.80 \to 2$, $<0.90 \to 3$, $\geq 0.90 \to 4$). Small-sample regime maps to \texttt{small\_data} via row count ($\geq 10{,}000 \to 0$, $\geq 5{,}000 \to 1$, $\geq 1{,}000 \to 2$, $\geq 500 \to 3$, $<500 \to 4$). A fifth dimension, \texttt{correlation\_handling}, is derived from a separate correlation detector (which counts the fraction of feature pairs with absolute Pearson correlation exceeding 0.7) that quantifies inter-feature dependency density. The sixth dimension, \texttt{privacy\_dp}, does not enter the requirement vector directly; instead, it influences scoring only through intent-conditioned scale factors (Section~\ref{sec:calibration}), allowing the privacy intent to upweight differential privacy capabilities without imposing them as a hard requirement for all datasets.

\subsection{Stress-Aware Capability Scoring}
\label{sec:candidate-selection}
SYNTHONY uses stress profiles to score and rank candidate model families \emph{without} running each generator. For each generator $g \in \mathcal{G}$, we maintain a capability registry $\mathbf{C}_g \in \{0,\ldots,4\}^J$ scoring the model's expected robustness along $J$ capability dimensions (e.g., handling of skewed distributions, high cardinality, small samples). Given a stress profile $\phi(D)$, SYNTHONY derives a requirement vector $\mathbf{r}(\phi(D))$ indicating which capabilities the dataset demands.

To account for the relative importance of capabilities under different intents, we introduce intent-conditioned scale factors $\boldsymbol{\alpha}_{i} =(\alpha_{i,1}, \ldots, \alpha_{i,J}) \in [0,10]^J$ for each intent $i \in \mathcal{I}$. The final score for generator $g$ under intent $i$ is:
\begin{equation}
    \text{Score}(g, i) = \sum_{j=1}^{J} \alpha_{i,j} \cdot c_{g,j} \cdot r_j(\phi(D))
    \label{eq:score}
\end{equation}
where $c_{g,j}$ is the capability score and $r_j(\phi(D))$ is the stress-derived requirement for dimension $j$. Generators are ranked by this score, and the top-ranked model is recommended.

The match function underlying Equation~\ref{eq:score} implements soft thresholding with graceful degradation: full credit (1.0) when the model's capability meets or exceeds the requirement, partial credit (0.7, 0.4) for near-misses, and zero otherwise. The full specification of the requirement derivation $f$ and match function $m_j$ is given in Appendix~\ref{sec:appendix-intent}.

\paragraph{Additive independence assumption.} The scoring function in Equation~\ref{eq:score} treats capability dimensions independently: each dimension contributes additively, so a model that is weak in one area is not penalized in another. This design enables interpretable, per-dimension scoring but may miss interaction effects---for instance, a dataset with \emph{both} high cardinality and severe skew may pose compounding challenges that exceed the sum of individual stresses. Capturing such interactions would require either higher-order terms in the scoring function or a learned model that implicitly discovers them from data.

\subsection{Calibration via Bayesian Optimization}
\label{sec:calibration}
Rather than relying solely on hand-crafted capability scores, we calibrate the entire scoring system against empirical benchmark data. We jointly optimize two parameter groups using Bayesian optimization: (1)~the capability scores $c_{g,j} \in \{0,\ldots,4\}$ for each generator $g$ and capability dimension $j$, and (2)~the intent-conditioned scale factors $\alpha_{i,j} \in [0,10]$ that weight each capability under intent $i$. For $|\mathcal{G}|=10$ generators, $|J|=6$ capabilities, and $|\mathcal{I}|=3$ intents, this yields $10 \times 6 + 3 \times 6 = 78$ parameters.
We maximize the average Spearman rank correlation between predicted and oracle rankings on the training split using Tree-structured Parzen Estimator (TPE) sampling \citep{bergstra2011algorithms} (500 trials, 50 startup).
Spearman correlation is chosen over Top-1 accuracy because it rewards correct ordering across the full ranking, not just the top pick.

Algorithm~\ref{alg:synthony} summarizes the full SYNTHONY selection pipeline.

\begin{algorithm}[t]
\caption{SYNTHONY Stress-Aware Selection}
\label{alg:synthony}
\begin{algorithmic}[1]
\REQUIRE Dataset $\mathcal{D}$, Intent $\mathcal{I}$, Calibrated registry $\mathbf{C}$, Scale factors $\boldsymbol{\alpha}$
\ENSURE Ranked generator list
\STATE Compute stress profile $\phi(\mathcal{D})$
\STATE Derive requirement vector $\mathbf{r} \leftarrow f(\phi(\mathcal{D}))$
\FOR{each generator $g \in \mathcal{G}$}
    \STATE $\text{Score}(g) \leftarrow \sum_{j} \alpha_{\mathcal{I},j} \cdot c_{g,j} \cdot r_j$
\ENDFOR
\RETURN Generators ranked by $\text{Score}(g)$ descending
\end{algorithmic}
\end{algorithm}

\section{Experiments}
\label{sec:experiments}

To validate the effectiveness of \textsc{Synthony} as a model selection agent, we conduct a systematic evaluation comparing its choices against intent-conditioned ground truth. Our goal is to answer: \textit{Can a stress-aware, intent-conditioned agent identify the optimal synthesizer for a given dataset and objective more reliably than heuristic rules, zero-shot LLMs, or meta-learning selectors?}

\subsection{Experimental Setup}

\paragraph{Benchmark Datasets}
We use 7 tabular datasets sourced from OpenML \citep{vanschoren2013openml} spanning diverse characteristics: Abalone, Bean, IndianLiverPatient, Obesity, faults, insurance, and wilt. These datasets vary in dimensionality ($6$--$17$ features), sample size ($579$--$13{,}611$ rows), and feature types (continuous, categorical, mixed), creating a robust test bed. We evaluate under three intents (privacy, fidelity, utility), yielding $7 \times 3 = 21$ dataset-intent pairs. We split at the dataset level (seed=42): 4 datasets for training and 3 for testing, producing $4 \times 3 = 12$ training pairs and $3 \times 3 = 9$ test pairs. Dataset-level splitting ensures that no dataset appears in both splits, preventing trivial memorization. All tables report held-out test set performance unless noted otherwise.

\paragraph{Candidate Generators}
For each dataset, we evaluate a pool of $N{=}10$ synthesizers spanning four categories: \textbf{Differential Privacy} (AIM, DPCART), \textbf{Deep Generative} (AutoDiff, TabDDPM, TVAE, NFlow), \textbf{Tree-Based} (ARF, CART), and \textbf{Statistical} (BayesianNetwork, SMOTE). We exclude the Identity (copy) baseline, as it trivially achieves perfect fidelity but zero privacy, making it uninformative for selection. One dataset-model pair (insurance, SMOTE) is absent, so insurance rankings contain 9 models while all others contain 10.
\paragraph{Ground Truth (Intent-Conditioned Oracles)}
To establish ground truth, we train every generator $g \in \mathcal{G}$ on every dataset $d \in \mathcal{D}_{\mathrm{bench}}$ and compute intent-specific metrics:
\textbf{Privacy} (Proportion Closer to Real, lower is better; for ranking we invert so that higher rank = better privacy),
\textbf{Fidelity} (Column Shape Score, higher is better), and
\textbf{Utility} (downstream ML task performance: test ROC AUC for classification datasets, test adjusted R$^2$ for regression datasets; higher is better).
Privacy and fidelity metrics are sourced from the benchmark's cleaned results. Utility scores are computed from a separate Tabular Utility evaluation, which trains classifiers/regressors on synthetic data and evaluates on held-out real test data.
For each dataset-intent pair, we rank all available models by the corresponding metric, producing an oracle ranking $g^*_d(\mathbf{w})$.

\subsection{Baselines}
We compare \textsc{Synthony} against four selection strategies.
\textbf{Static Heuristic:} a logic-based baseline mimicking practitioner heuristics using only row count and intent (tree-based for utility/fidelity, DP models for privacy), without dataset profiling beyond row count.
\textbf{Vanilla LLM Selector:} a GPT-4o-mini zero-shot selector given the dataset schema and summary statistics but no stress profiles or capability scores (temperature~0).
\textbf{Random Search:} uniform random generator sampling (expected value over 1{,}000 trials).
\textbf{Meta-feature $k$NN Selector:} a meta-learning baseline extracting 9 dataset meta-features, finding $k{=}3$ nearest training datasets by Euclidean distance, and aggregating their oracle rankings. Because it interpolates from oracle rankings of training datasets, $k$NN has a structural advantage over zero-knowledge methods: it leverages ground-truth evaluation data at training time. We report it above the midrule in Table~\ref{tab:main} as an \emph{oracle-informed reference point} rather than a deployable zero-knowledge method.

\subsection{Evaluation Metrics}
We assess selection quality with four metrics computed over the 10-model rankings. Our two primary metrics are \textbf{Top-3 Accuracy}, the fraction of dataset-intent pairs where the oracle-best model appears in the agent's top-3 recommendations (reflecting the realistic deployment scenario of presenting a short list for downstream evaluation), and \textbf{Spearman Rank Correlation}, which measures agreement between predicted and oracle full rankings across all 10 models ($[-1, 1]$) and rewards correct ordering across the entire ranking. As secondary metrics we report \textbf{Top-1 Accuracy}, the fraction of pairs where the selector's top pick matches the oracle-best model (with only 9 test pairs, each correct prediction shifts Top-1 by 11.1 percentage points, so we interpret differences smaller than ${\sim}$0.15 with caution), and \textbf{NDCG} (Normalized Discounted Cumulative Gain), which evaluates the full predicted ranking with position-weighted relevance ($[0, 1]$).

\subsection{SYNTHONY Optimization Variants}
We evaluate two optimization strategies. \textbf{SF-only} tunes 18 scale factor parameters ($6$ capabilities $\times$ $3$ intents) via Bayesian optimization (Optuna, 200 trials, TPE sampler), holding hand-crafted capability scores fixed, with Top-1 accuracy as the training objective. \textbf{Joint optimization} simultaneously tunes 78 parameters (60 capability scores, $|\mathcal{G}|{=}10 \times |J|{=}6$, integer $\{0,\ldots,4\}$; plus 18 scale factors, float $[0,10]$) over 500 trials, optimizing average Spearman on the training set. We use Spearman for the larger search space because Top-1 is too sparse a signal for 78 parameters. These variants differ in both parameter count and objective, so performance differences reflect both changes; controlled variants appear in Appendix~\ref{sec:appendix-optimization}.

\subsection{Results}

\paragraph{Selection Performance}

Tabular synthesizers exhibit highly non-uniform performance across datasets and intents. Privacy exhibits genuine dataset-dependent variation: the best model varies across datasets, with multiple model families each winning on different datasets. Fidelity and utility also show dataset-dependent patterns, though with somewhat stronger concentration among tree-based and statistical models. This motivates intent-conditioned selection: no single model can serve all objectives.

Table~\ref{tab:main} summarizes test set performance across 9 held-out dataset-intent pairs. We highlight Top-3 accuracy and Spearman correlation as the primary comparison metrics.

\begin{table}[t]
\centering
\caption{Test set performance of intent-conditioned model selection strategies (9 dataset-intent pairs). Primary metrics are \textit{italicized}. $k$NN$^\dagger$ uses oracle neighbor rankings from training datasets and serves as an \emph{oracle-informed reference point}, not a deployable zero-knowledge method. Bold indicates best among zero-knowledge methods. With 9 test pairs, each Top-1 prediction shifts accuracy by ${\approx}$11.1 percentage points.}
\label{tab:main}
\begin{tabular}{lcccc}
\toprule
Strategy & Top-1 & \textit{Top-3} & \textit{Spear.} & NDCG \\
\midrule
$k$NN (k=3)$^\dagger$ \textit{(oracle-informed)} & .556 & \textit{.778} & \textit{.578} & .918 \\
\midrule
Random (E[1000]) & .111 & \textit{.304} & \textit{$-.001$} & .815 \\
Vanilla LLM (GPT-4o-mini) & .000 & \textit{.111} & \textit{$-.087$} & .784 \\
Static Heuristic & .333 & \textit{.778} & \textit{.422} & .889 \\
\textsc{Synthony} (joint opt) & .000 & \textit{.556} & \textit{.557} & .893 \\
\textsc{Synthony} (SF-only) & \textbf{.444} & \textit{\textbf{.667}} & \textit{\textbf{.573}} & \textbf{.914} \\
\bottomrule
\end{tabular}
\end{table}

Among zero-knowledge methods, \textsc{Synthony} (SF-only) achieves the best Spearman ($\rho = 0.573$) and Top-1 ($0.444$), substantially outperforming the Static Heuristic ($0.422$ Spearman) and the Vanilla LLM ($-0.087$, worse than random). The SF-only variant nearly matches the oracle-informed $k$NN reference ($0.578$ Spearman) despite a fundamental asymmetry: $k$NN interpolates from ground-truth rankings of training datasets, while \textsc{Synthony} requires \emph{no} oracle scores at inference time. This makes \textsc{Synthony} applicable to new synthesizers added to the registry without rerunning benchmarks---a practical advantage $k$NN cannot offer. Furthermore, \textsc{Synthony}'s capability-based scoring provides interpretable, per-dimension explanations for each recommendation, whereas $k$NN's neighbor-aggregation is opaque about \emph{why} a model is suitable.

The joint optimization does not generalize as well: test Spearman ($0.557$) is slightly below SF-only, and Top-1 drops to $0.000$. This suggests overfitting when optimizing 78 parameters on 12 training pairs, though the comparison is confounded by different training objectives (Top-1 vs.\ Spearman). The Static Heuristic achieves high Top-3 ($77.8\%$) by placing strong models within the top~3, but its poor Spearman ($0.422$) indicates weak full-ranking quality.

\paragraph{Per-Intent Breakdown}
To disentangle selection quality from CART's frequent wins (8/21 oracle-best across all pairs), we examine per-intent oracle patterns across all 7 benchmark datasets, not just 3 test dataset. Under \textbf{fidelity}, CART wins 4/7 datasets; under \textbf{utility}, CART wins 4/7 while SMOTE takes the remaining 3. The discriminative test is \textbf{privacy}, where five distinct models win across seven datasets (TabDDPM 2, AutoDiff 2, AIM/DPCART/BayesianNetwork each 1). A selector that simply always predicts CART would achieve $8/21 = 38\%$ Top-1 overall but $0\%$ on privacy. \textsc{Synthony}'s stress-aware scoring provides genuine signal beyond predicting the majority class: it differentiates utility from fidelity requirements and identifies non-CART winners on datasets where tree-based models underperform.

On the 9 test pairs specifically, \textsc{Synthony} (SF-only) correctly identifies the oracle-best model on 4/9 pairs (Top-1 = $0.444$). Its errors concentrate on the privacy intent, the hardest for all methods---consistent with the diversity of oracle winners under privacy across the full benchmark. The Static Heuristic achieves 3/9 Top-1, but all its correct predictions fall under fidelity and utility where CART dominates; it achieves 0\% Top-1 on privacy. \textsc{Synthony}'s correct privacy predictions, though limited, demonstrate that stress profiling captures signals that simple heuristics miss.

\paragraph{CART Dominance and Benchmark Scope}
CART's strong showing (8/21 oracle-best overall) partly reflects the composition of our 7-dataset benchmark: most datasets are moderate-sized ($500$--$13{,}000$ rows) with relatively low cardinality, a regime where tree-based methods are known to excel. We note this is consistent with recent independent benchmarks \citep{dsj2025benchmarking,kindji2025tabular} that also find tree-based methods competitive in this data range. Expanding to datasets with higher cardinality, more features, or larger scale (${>}100{,}000$ rows) would likely shift the distribution of oracle winners toward deep generative models and reduce single-model dominance.

\paragraph{Privacy Intent: Challenges and Signal}
The privacy intent presents a circular challenge: it is where the discriminative signal for model selection is strongest (five distinct oracle winners across seven datasets), but it is also the hardest to model because no single architectural property reliably predicts privacy performance. DP-certified models (AIM, DPCART) do not always achieve the best privacy scores, as non-DP models like TabDDPM can achieve lower re-identification risk through smoother density estimation. This makes privacy the most compelling use case for stress-aware selection---and the one requiring the most training data to calibrate reliably.


\subsection{Ablation Study}

To isolate the contribution of each \textsc{Synthony} component, we systematically disable them (Table~\ref{tab:ablation}). \textsc{Synthony}'s SF-only pipeline consists of three components: (A)~stress profiling (detecting required capabilities from dataset statistics), (B)~capability matching (scoring models against requirements using the capability registry), and (C)~focus-based scaling (Bayesian-optimized scale factors per intent). All ablation variants use the SF-only optimization (18 parameters) to isolate component-level effects.

\begin{table}[t]
\centering
\caption{Ablation study on the SF-only pipeline (test set, 9 pairs). Components: (A)~stress profiling, (B)~capability matching, (C)~focus scaling. Stress profiling provides the strongest signal; adding focus scaling degrades test performance due to overfitting at this sample size ($n{=}12$ training pairs, 18 parameters).}
\label{tab:ablation}
\begin{tabular}{lcccc}
\toprule
Variant & Top-1 & Top-3 & Spear. & NDCG \\
\midrule
Vanilla LLM (no \textsc{Synthony}) & .000 & .111 & $-.087$ & .784 \\
$-$ Stress $-$ Focus (bare scoring) & .444 & .667 & .478 & .904 \\
$-$ Focus scaling (stress only) & .444 & .667 & .602 & .910 \\
$-$ Stress profiling (focus only) & .333 & .667 & .430 & .901 \\
Full \textsc{Synthony} (SF-only) & .444 & .667 & .573 & .914 \\
\bottomrule
\end{tabular}
\end{table}

Table~\ref{tab:ablation} reveals an interesting pattern. Removing focus scaling (``$-$ Focus scaling'') actually \emph{improves} Spearman from $0.573$ to $0.602$, suggesting that the optimized scale factors slightly overfit on 12 training pairs. The stress-only variant achieves the best Spearman among all ablations while maintaining Top-1 ($0.444$) and Top-3 ($0.667$) accuracy. This does not invalidate intent-based scaling as a concept, but indicates that 12 training pairs are insufficient to learn 18 scale parameters reliably. We conjecture that with a larger benchmark ($\geq$20 datasets), focus scaling would become beneficial---particularly for the privacy intent, where intent-specific weighting should upweight DP capability for privacy-focused users.

Removing stress profiling (``$-$ Stress profiling'') causes the largest Spearman drop ($0.573 \to 0.430$), confirming that dataset-derived capability requirements are the most important signal. Without stress profiling, the system cannot differentiate which models suit which datasets; the remaining score differences come only from intent-conditioned scale factors applied to raw capability scores.

The bare scoring variant (``$-$ Stress $-$ Focus'') still achieves $0.478$ Spearman, showing that even unweighted capability matching provides signal above random ($-0.001$). The full system combines both components to achieve the best overall balance across all four metrics.

\subsection{Do Stress Profiles Predict Synthesizer Families?}

\begin{table}[t]
\centering
\caption{Predicting the best synthesizer from dataset meta-features (test set, 9 pairs). Majority vote's $33.3\%$ Top-1 reflects CART's frequent wins (8/21 oracle-best); the discriminative signal is in Spearman, where $k$NN ($0.578$) substantially outperforms frequency-based and parametric baselines ($0.293$--$0.300$).}
\label{tab:stress-predict}
\begin{tabular}{lcccc}
\toprule
Predictor & Top-1 & Top-3 & Spear. & NDCG \\
\midrule
Majority vote & .333 & .778 & .300 & .899 \\
Logistic regression & .556 & .889 & .293 & .893 \\
Decision tree & .222 & .667 & .299 & .900 \\
$k$NN (k=3) & .556 & .778 & .578 & .918 \\
\bottomrule
\end{tabular}
\end{table}

Table~\ref{tab:stress-predict} validates that dataset meta-features (including stress profile dimensions) are predictive of which synthesizer succeeds. Logistic regression achieves the highest Top-1 ($0.556$) and Top-3 ($0.889$) accuracy, but $k$NN ($k{=}3$) achieves the best Spearman correlation ($0.578$) by nearly double, demonstrating that meta-feature similarity produces better \emph{full rankings}, not just correct top-1 predictions. This supports the stress profiling hypothesis: dataset difficulty signatures predict synthesizer efficacy in a systematic, learnable way.

\paragraph{Implications for SYNTHONY.} Remarkably, \textsc{Synthony} (SF-only) nearly matches the $k$NN stress predictor on Spearman ($0.573$ vs.\ $0.578$) despite using a fundamentally different approach: capability matching rather than neighbor-based interpolation. The joint optimization does not improve on this, suggesting that the hand-crafted capability scores already encode useful inductive bias that is lost when all 78 parameters are freely optimized. The remaining gap with $k$NN reflects a structural difference: $k$NN directly interpolates oracle rankings from similar datasets, while \textsc{Synthony} must generalize through a capability-based abstraction layer. Crucially, this near-parity is achieved without access to oracle rankings at test time---\textsc{Synthony}'s capability registry generalizes to new datasets without rerunning any synthesizer, while $k$NN requires precomputed rankings for all training datasets.

\paragraph{Analysis of ``Hard'' Datasets}
The privacy intent is the most challenging: no single model dominates, and the best choice varies across datasets (TabDDPM wins 2/7, AutoDiff wins 2/7, AIM/DPCART/BayesianNetwork each win 1/7). All methods exhibit the lowest accuracy on privacy, where the Static Heuristic assumption that DP-certified models (DPCART, AIM) always minimize re-identification risk is violated: on datasets with moderate dimensionality, non-DP diffusion models (TabDDPM or AutoDiff) achieve lower Distance-to-Closest-Record scores by learning smoother density estimates that avoid near-copying. The $k$NN selector handles privacy best by aggregating neighbor-specific patterns rather than relying on a fixed rule. Utility presents an interesting challenge: CART wins 4/7 datasets but SMOTE takes the other 3, and the best synthesizer depends on both dataset characteristics and the downstream task type (classification vs.\ regression), requiring selectors to account for this interaction.

\section{Conclusion}
\label{sec:conclusion}

We studied intent-conditioned model selection for tabular data synthesis and proposed \textbf{stress profiling}, a set of interpretable meta-features capturing dataset difficulty, integrated into \textsc{SYNTHONY}, a selection framework that matches stress profiles against a calibrated capability registry.

Our evaluation yields three main findings. First, \textbf{stress profiling is predictive}: a $k$NN meta-learner achieves 77.8\% Top-3 and 55.6\% Top-1 on held-out pairs (Table~\ref{tab:stress-predict}), confirming that DGM failures are systematic and predictable from dataset characteristics. Second, \textbf{capability matching nearly reaches oracle-informed meta-learning}: \textsc{Synthony} (SF-only) achieves $0.573$ Spearman, nearly matching $k$NN ($0.578$), while offering practical advantages that $k$NN lacks---zero-shot applicability to new synthesizers (no benchmark rerunning needed), interpretable per-dimension explanations for each recommendation, and the ability to incorporate domain knowledge through the capability registry. The joint optimization ($0.557$ Spearman) does not improve on the test set, indicating that the hand-crafted capability scores encode useful inductive bias lost when all 78 parameters are freely optimized on 12 training pairs. Third, \textbf{zero-shot LLMs are insufficient}: GPT-4o-mini achieves 0\% Top-1 and $-0.087$ Spearman, confirming that synthesis-specific calibration is essential.

\paragraph{Limitations.} Our 7-dataset, 9-pair test set limits statistical power. CART wins 4/7 fidelity and 4/7 utility pairs, making the problem partially solvable by majority-class prediction (33.3\% Top-1). The SF-only and joint variants differ in both parameter space and objective (Top-1 vs.\ Spearman), partially confounding the comparison; controlled variants appear in Appendix~\ref{sec:appendix-optimization}.

\paragraph{Future Work.} The primary direction is \emph{learning} capability scores from data rather than hand-crafting them: replacing the static registry with learned embeddings~\citep{vanschoren2019meta}, incorporating stress meta-features directly into the scoring function, and expanding to ${\geq}20$ datasets and additional model families (e.g., GReaT~\citep{borisov2022great}) to reduce single-model dominance effects.

\subsubsection*{Reproducibility Statement}
Code, data, and the capability registry are publicly available at \url{https://github.com/UCLA-Trustworthy-AI-Lab/Synthony}.

\bibliography{iclr2026_conference}
\bibliographystyle{iclr2026_delta}

\newpage
\appendix

\section{Joint Optimization of Capability Scores and Scale Factors}
\label{sec:appendix-optimization}

This appendix provides full details of the Bayesian optimization procedure used to calibrate \textsc{SYNTHONY}'s scoring system against empirical benchmark data.

\subsection{Motivation}

\textsc{SYNTHONY}'s recommendation engine scores candidate synthesizers by matching dataset stress profiles against a capability registry. Each model $g$ has a capability vector $\mathbf{c}_g \in \{0,\ldots,4\}^6$ across six dimensions: skew handling, cardinality handling, Zipfian handling, small data, correlation handling, and differential privacy. Under a given intent $i$, scale factors $\boldsymbol{\alpha}_i \in [0,10]^6$ weight each capability dimension, so the total score for model $g$ on dataset $d$ under intent $i$ is:
\begin{equation}
    S(g, d, i) = \sum_{j=1}^{6} \alpha_{i,j} \cdot w_j(d) \cdot m_j(c_{g,j}, r_j(d))
\end{equation}
where $r_j(d)$ is the required capability level derived from the dataset stress profile, $w_j(d)$ is a base weight (1.0 if capability $j$ is required, 0.1 otherwise), and $m_j$ is a match function that compares $c_{g,j}$ against $r_j(d)$.

The initial capability scores were assigned by human judgment based on model architecture properties and the literature. Before calibration (i.e., with uniform scale factors), the SF-only variant with hand-crafted capabilities achieves 0.444 test Top-1 and 0.573 Spearman (Table~\ref{tab:main}), showing that the hand-crafted scores already encode useful inductive bias but leave room for improvement through calibration.

\subsection{Parameter Space}

We jointly optimize two parameter groups:

\paragraph{Capability scores (60 parameters).} For each of the 10 overlap models (AIM, ARF, AutoDiff, BayesianNetwork, CART, DPCART, NFlow, SMOTE, TabDDPM, TVAE) and 6 capability dimensions, we optimize $c_{g,j} \in \{0, 1, 2, 3, 4\}$ as integer parameters.

\paragraph{Scale factors (18 parameters).} For each of the 3 intents (privacy, fidelity, utility) and 6 capability dimensions, we optimize $\alpha_{i,j} \in [0.0, 10.0]$ as continuous parameters.

This yields a total of $10 \times 6 + 3 \times 6 = 78$ parameters.

\subsection{Objective Function}

We maximize the average Spearman rank correlation between predicted and oracle rankings on the training split (12 dataset--intent pairs):
\begin{equation}
    \max_{\mathbf{c}, \boldsymbol{\alpha}} \; \frac{1}{|\mathcal{T}_{\text{train}}|} \sum_{(d,i) \in \mathcal{T}_{\text{train}}} \rho\bigl(\hat{\sigma}(d,i;\mathbf{c},\boldsymbol{\alpha}),\; \sigma^*(d,i)\bigr)
\end{equation}
where $\hat{\sigma}$ is the predicted ranking (sorted by $S(g,d,i)$), $\sigma^*$ is the oracle ranking from ground truth, and $\rho$ is Spearman's rank correlation. Only models present in both rankings contribute to $\rho$.

Spearman correlation was chosen over Top-1 accuracy because it provides a denser gradient signal: Top-1 is a binary 0/1 reward that does not distinguish between second place and last place, while Spearman rewards getting the full ordering correct.

\subsection{Optimization Configuration}

We use the Tree-structured Parzen Estimator (TPE) sampler \citep{bergstra2011algorithms} implemented in Optuna:
\begin{itemize}
    \item \textbf{Trials:} 500 (50 random startup trials, 450 TPE-guided)
    \item \textbf{Seed:} 42 (reproducible)
    \item \textbf{Direction:} maximize
    \item \textbf{Runtime:} ${\sim}35$ seconds on a single CPU core
\end{itemize}

During each trial, the engine's capability registry is temporarily overwritten with the sampled values. The engine then runs its standard scoring pipeline (hard filters $\to$ stress profiling $\to$ capability matching $\to$ scale-factor-weighted scoring) on all training pairs.

\subsection{Results}

Table~\ref{tab:joint-opt-summary} summarizes performance on both splits.

\begin{table}[h]
\centering
\caption{Joint optimization results on train and test splits.}
\label{tab:joint-opt-summary}
\begin{tabular}{lcccc}
\toprule
Split & Top-1 & Top-3 & Spearman & NDCG \\
\midrule
Training (12 pairs) & .250 & .750 & .610 & .897 \\
Test (9 pairs) & .000 & .556 & .557 & .893 \\
\bottomrule
\end{tabular}
\end{table}

\subsection{Optimized Capability Scores}

Table~\ref{tab:opt-caps} shows the learned capability scores alongside the original hand-crafted values for all 10 models.

\begin{table}[h]
\centering
\caption{Capability scores: hand-crafted (left) vs.\ optimized (right). Changes of $\geq2$ are bolded.}
\label{tab:opt-caps}
\resizebox{\linewidth}{!}{%
\begin{tabular}{l*{6}{cc}}
\toprule
 & \multicolumn{2}{c}{Skew} & \multicolumn{2}{c}{Cardinal.} & \multicolumn{2}{c}{Zipfian} & \multicolumn{2}{c}{Small data} & \multicolumn{2}{c}{Correl.} & \multicolumn{2}{c}{Privacy} \\
\cmidrule(lr){2-3}\cmidrule(lr){4-5}\cmidrule(lr){6-7}\cmidrule(lr){8-9}\cmidrule(lr){10-11}\cmidrule(lr){12-13}
Model & Orig & Opt & Orig & Opt & Orig & Opt & Orig & Opt & Orig & Opt & Orig & Opt \\
\midrule
AIM     & 3 & 2 & 0 & 0 & 1 & \textbf{3} & 2 & \textbf{4} & 3 & \textbf{0} & 4 & \textbf{2} \\
ARF     & 2 & \textbf{4} & 4 & \textbf{2} & 3 & 4 & 4 & \textbf{0} & 4 & \textbf{1} & 0 & 1 \\
AutoDiff & 1 & \textbf{4} & 3 & \textbf{1} & 2 & 3 & 2 & \textbf{0} & 1 & \textbf{3} & 0 & \textbf{4} \\
BayesianNet. & 3 & 4 & 4 & 4 & 2 & 2 & 4 & 4 & 3 & 2 & 0 & \textbf{4} \\
CART    & 3 & 4 & 4 & \textbf{2} & 2 & 2 & 4 & 4 & 4 & 3 & 0 & \textbf{4} \\
DPCART  & 2 & 3 & 0 & 0 & 2 & \textbf{4} & 2 & 1 & 3 & \textbf{0} & 3 & \textbf{0} \\
NFlow   & 2 & \textbf{4} & 4 & \textbf{2} & 2 & \textbf{0} & 4 & \textbf{0} & 1 & 1 & 0 & 0 \\
SMOTE   & 3 & 2 & 4 & \textbf{2} & 2 & 2 & 4 & \textbf{1} & 4 & 4 & 0 & \textbf{3} \\
TabDDPM & 1 & 2 & 2 & 2 & 2 & \textbf{0} & 2 & \textbf{0} & 3 & 2 & 0 & \textbf{2} \\
TVAE    & 2 & \textbf{0} & 4 & \textbf{2} & 1 & 2 & 3 & 2 & 4 & \textbf{2} & 0 & \textbf{3} \\
\bottomrule
\end{tabular}%
}
\end{table}

\subsection{Optimized Scale Factors}

Table~\ref{tab:opt-sf} shows the learned intent-conditioned scale factors.

\begin{table}[h]
\centering
\caption{Optimized scale factors $\alpha_{i,j}$ per intent and capability dimension.}
\label{tab:opt-sf}
\begin{tabular}{lcccccc}
\toprule
Intent & Skew & Cardinal. & Zipfian & Small & Correl. & Privacy \\
\midrule
Privacy  & 5.69 & 3.78 & 7.32 & 9.83 & 8.90 & 9.10 \\
Fidelity & 2.43 & 7.36 & 4.10 & 3.11 & 2.78 & 9.44 \\
Utility  & 0.68 & 2.59 & 3.71 & 5.36 & 1.42 & 2.96 \\
\bottomrule
\end{tabular}
\end{table}

\paragraph{Interpretation caveat.} Some scale factor values reflect spurious correlations in the small training set rather than meaningful intent-capability relationships. For example, the fidelity intent assigns its highest weight ($9.44$) to the privacy/DP dimension. This occurs because models that happen to score well on fidelity in the training data (e.g., CART, BayesianNetwork) also receive high optimized privacy capability scores, creating a coincidental correlation that the optimizer exploits. Such patterns are unlikely to generalize to benchmarks with different model pools or larger dataset collections. The scale factors should be interpreted as dataset-specific optimization artifacts rather than universal intent--capability mappings.

\section{Intent-Aware Scoring Design}
\label{sec:appendix-intent}

The intent-conditioned scoring mechanism in \textsc{SYNTHONY} operates through multiplicative scale factors that reweight the base capability matching scores. This section details the design rationale.

\subsection{Base Scoring}

For each eligible model $g$ and required capability $j$, the base match score is:
\[
m_j(c_{g,j}, r_j) = \begin{cases}
1.0 & \text{if } c_{g,j} \geq r_j \\
0.7 & \text{if } c_{g,j} = r_j - 1 \\
0.4 & \text{if } c_{g,j} = r_j - 2 \\
0.0 & \text{otherwise}
\end{cases}
\]
For non-required capabilities (where $r_j = 0$), when scale factors are active, the match score is $c_{g,j}/4$ to differentiate models by raw capability strength.

\subsection{Scale Factor Application}

Under an intent $i$, the weighted contribution of capability $j$ to model $g$'s total score is:
\[
\text{contribution}_j = m_j \cdot w_j \cdot \alpha_{i,j}
\]
where $w_j = 1.0$ if the dataset requires capability $j$ (i.e., $r_j > 0$) and $w_j = 0.1$ otherwise. The total score is $S(g) = \sum_j \text{contribution}_j$.

\subsection{Design Decisions}

\paragraph{Bypassing the hard-problem path.} When scale factors are provided, the engine's ``hard problem'' routing (which forces selection of GReaT for datasets with simultaneous skew + cardinality + Zipfian stress) is bypassed. This allows the scale-factor-weighted scoring to make the selection decision, which is essential for the optimizer to explore the full scoring space.

\paragraph{Bypassing tie-breaking.} Similarly, the heuristic tie-breaking rules (e.g., ``prefer ARF for small data'') are skipped when scale factors are active, since the scale factors already encode user preference through learned weights.

\paragraph{Integer capability scores.} We constrain capability scores to integers $\{0,\ldots,4\}$ rather than continuous values. This preserves interpretability (scores map to qualitative levels: 0=none, 1=poor, 2=moderate, 3=good, 4=excellent) and reduces the effective search space, improving sample efficiency of the Bayesian optimizer.

\section{System Architecture}
\label{sec:appendix-architecture}

This appendix describes \textsc{SYNTHONY}'s implementation as a deployable system,
complementing the algorithmic description in the main text
(Sections~\ref{sec:method}--\ref{sec:experiments}).

\subsection{System Overview}
\label{sec:arch-overview}

\begin{figure}[ht]
\centering
\begin{tikzpicture}[
    node distance=0.8cm and 1.2cm,
    box/.style={draw, rounded corners, minimum width=2.6cm, minimum height=0.9cm,
                align=center, font=\small},
    pkg/.style={draw, dashed, rounded corners, inner sep=6pt, fill=#1!8},
    arr/.style={-{Stealth[length=2.5mm]}, thick},
    darr/.style={-{Stealth[length=2.5mm]}, thick, dashed},
    lbl/.style={font=\footnotesize\itshape, text=gray!70!black},
]

\node[box, fill=pink!30] (user) {User / Agent};

\node[box, fill=blue!15, below=1.0cm of user] (api) {API Gateway\\[-2pt]{\scriptsize FastAPI / MCP / CLI}};

\node[box, fill=green!15, below left=1.8cm and 3.2cm of api] (profiler) {Stochastic Data\\[-2pt]Analyzer};
\node[box, fill=green!15, below=0.5cm of profiler] (colprof) {Column\\[-2pt]Analyzer};

\node[box, fill=green!15, below=3.0cm of api] (engine) {Recommender\\[-2pt]Engine};
\node[box, fill=green!15, below=0.5cm of engine] (scorer) {Hybrid\\[-2pt]Scorer};

\node[box, fill=yellow!20, right=2.5cm of engine] (registry) {Capability\\[-2pt]Registry};

\node[box, fill=orange!15, above=1.5cm of registry] (bench) {Benchmark\\[-2pt]Runner};

\begin{scope}[on background layer]
    \node[pkg=green, fit=(profiler)(colprof),
          label={[lbl]above:{\textbf{1\,\textperiodcentered\,}Data Infrastructure}}] (pkg1) {};
    \node[pkg=yellow, fit=(registry),
          label={[lbl]below:{\textbf{2\,\textperiodcentered\,}Model Core}}] {};
    \node[pkg=green, fit=(engine)(scorer),
          label={[lbl]above:{\textbf{3\,\textperiodcentered\,}Orchestration}}] {};
    \node[pkg=orange, fit=(bench),
          label={[lbl]above:Offline}] {};
\end{scope}


\draw[arr] (user) -- node[right, lbl] {CSV / intent} (api);

\draw[arr] (api.west) -- ++(-1.6,0) |- node[right, lbl, pos=0.3] {raw data}(profiler.east);
\draw[arr] (api.west) ++(-1.6,0) |- (colprof.east);

\draw[arr] (profiler.east) -- ++(2.0,0) |- node[above, lbl, pos=0.7] {$\phi(D)$} (engine.west);
\draw[arr] (colprof.east) -- ++(2.0,0) |- (engine.south west);

\draw[arr] (registry.west) -- node[above, lbl, pos =0.3] {$\mathbf{C}_g$} (engine.east);

\draw[arr] (engine) -- (scorer);

\draw[arr] (scorer.east) -- ++(1.4,0) |- node[right, lbl, pos=0.25] {ranking} (api.east);

\draw[arr] (api.north east) -- ++(0.5, 0.5) -- node[right, lbl] {JSON} (user.south east);

\draw[darr] (bench) -- node[right, lbl] {calibrate} (registry);

\end{tikzpicture}
\caption{SYNTHONY system architecture. Solid arrows denote the real-time
  recommendation pipeline; dashed arrows denote the offline calibration loop
  (Section~\ref{sec:calibration}). Data flows left through profiling, center
  through scoring, and returns right through the API.}
\label{fig:arch-overview}
\end{figure}
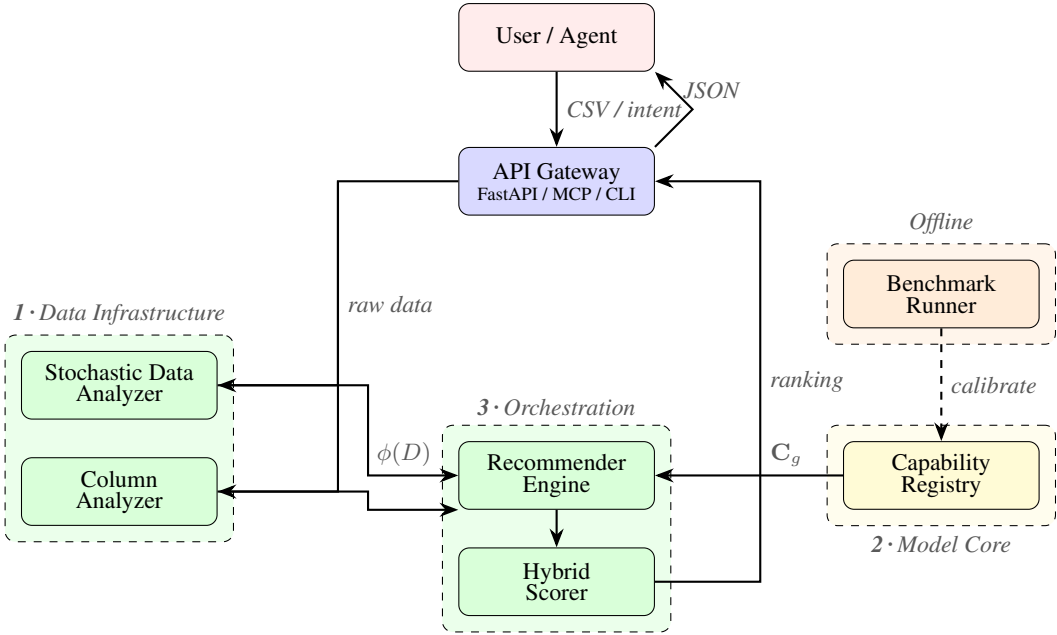

\textsc{SYNTHONY} is organized into three packages that form a data-processing
pipeline (Figure~\ref{fig:arch-overview}):
(1)~\textbf{Data Infrastructure} profiles raw datasets into stress vectors,
(2)~\textbf{Model Core} maintains the capability registry and scoring logic,
and (3)~\textbf{Orchestration} combines profiles with registry scores to
produce ranked recommendations. Three interface layers---a REST API, an MCP
server, and a CLI---expose this pipeline to different consumer types (human
users, AI agents, and scripts, respectively).

\paragraph{Package 1: Data Infrastructure.}
The \texttt{StochasticDataAnalyzer} applies four detectors---skewness,
cardinality, Zipfian concentration, and data-size classification---to produce
a stress profile $\phi(D) \in \mathbb{R}^4$ (Section~\ref{sec:stress-profiling}). The \texttt{ColumnAnalyzer} computes per-column difficulty scores (type inference, missing-value density, cardinality ratio) that provide fine-grained context for the LLM-based recommendation path.

\paragraph{Package 2: Model Core.}
The capability registry (\texttt{model\_capabilities.json}, v7.0.0) stores 15 synthesizer specifications, each containing: (i)~capability scores $\mathbf{c}_g \in \{0,\ldots,4\}^6$ across six dimensions, (ii)~empirical quality metrics from Spark benchmarks (10 datasets, 14 models), (iii)~constraint metadata (GPU requirements, min/max row limits, DP certification), and (iv)~engine configuration (score decay curve, tie-breaking priorities, hard-problem routing).

\paragraph{Package 3: Orchestration.}
The \texttt{ModelRecommendationEngine} implements the full selection pipeline
described in Algorithm~\ref{alg:synthony}: hard filtering $\to$ stress
profiling $\to$ capability matching $\to$ scale-factor-weighted scoring $\to$
tie-breaking. Three recommendation methods are supported: \texttt{rule\_based}
(deterministic, sub-second), \texttt{llm} (OpenAI/vLLM-backed, 5--30s), and
\texttt{hybrid} (rule-based candidates re-ranked by LLM).

%
\subsection{Recommendation Scoring Pipeline}
\label{sec:appendix-rs-logic}

\begin{figure}[ht]
\centering
\begin{tikzpicture}[
    node distance=0.7cm,
    flowstep/.style={draw, rounded corners, minimum width=4.5cm, minimum height=0.7cm,
                 align=center, font=\small, fill=green!10},
    decision/.style={draw, diamond, aspect=2.5, minimum width=2cm,
                     align=center, font=\small, fill=yellow!15},
    exit/.style={draw, rounded corners, minimum width=3cm, minimum height=0.7cm,
                 align=center, font=\small, fill=red!10},
    result/.style={draw, rounded corners, minimum width=4.5cm, minimum height=0.7cm,
                   align=center, font=\small, fill=blue!10},
    arr/.style={-{Stealth[length=2.5mm]}, thick},
    lbl/.style={font=\scriptsize},
]

\node[flowstep] (input) {Input: Dataset $D$, Intent $\mathcal{I}$};
\node[flowstep, below=of input] (filter) {1. Hard Filters\\[-2pt]{\scriptsize CPU-only, DP, row limits, exclusions}};
\node[decision, below=0.8cm of filter] (empty) {Pool\\empty?};
\node[flowstep, below=0.8cm of empty] (stress) {2. Compute Stress Profile $\phi(D)$};
\node[decision, below=0.8cm of stress] (hard) {Hard\\problem?};
\node[flowstep, below=0.8cm of hard] (reqcap) {3. Derive Requirements $\mathbf{r}(\phi)$};
\node[flowstep, below=of reqcap] (score) {4. Score: $\sum_j \alpha_{i,j} \cdot m_j(c_{g,j}, r_j)$\\[-2pt]{\scriptsize + empirical quality bonus}};
\node[flowstep, below=of score] (rank) {5. Rank by Score (descending)};
\node[decision, below=0.8cm of rank] (tie) {Top-2\\within 5\%?};
\node[flowstep, below=0.8cm of tie] (tiebreak) {6. Tie-Break\\[-2pt]{\scriptsize Small data $\to$ ARF; Speed $\to$ CART}};
\node[result, below=of tiebreak] (output) {7. Return Ranked List + Confidence};

\node[exit, right=2cm of empty] (nomodel) {Error: No eligible\\models};
\node[exit, right=2cm of hard] (hardroute) {Hard Problem Route\\[-2pt]{\scriptsize GReaT $\to$ fallback list}};

\draw[arr] (input) -- (filter);
\draw[arr] (filter) -- (empty);
\draw[arr] (empty) -- node[left, lbl] {no} (stress);
\draw[arr] (empty) -- node[above, lbl] {yes} (nomodel);
\draw[arr] (stress) -- (hard);
\draw[arr] (hard) -- node[left, lbl] {no} (reqcap);
\draw[arr] (hard) -- node[above, lbl] {yes} (hardroute);
\draw[arr] (reqcap) -- (score);
\draw[arr] (score) -- (rank);
\draw[arr] (rank) -- (tie);
\draw[arr] (tie) -- node[left, lbl] {yes} (tiebreak);
\draw[arr] (tie.east) -- ++(1.5,0) |- node[right, lbl, pos=0.25] {no} (output.east);
\draw[arr] (tiebreak) -- (output);

\end{tikzpicture}
\caption{Rule-based scoring pipeline. The match function $m_j$ implements a decay curve: 1.0 (exact match), 0.7 (one level below), 0.4 (two levels below), 0.0 (otherwise). Scale factors $\alpha_{i,j}$ are applied when an intent is specified. The hard-problem path is bypassed when scale factors are provided to allow the optimizer full control.}
\label{fig:rs-pipeline}
\end{figure}
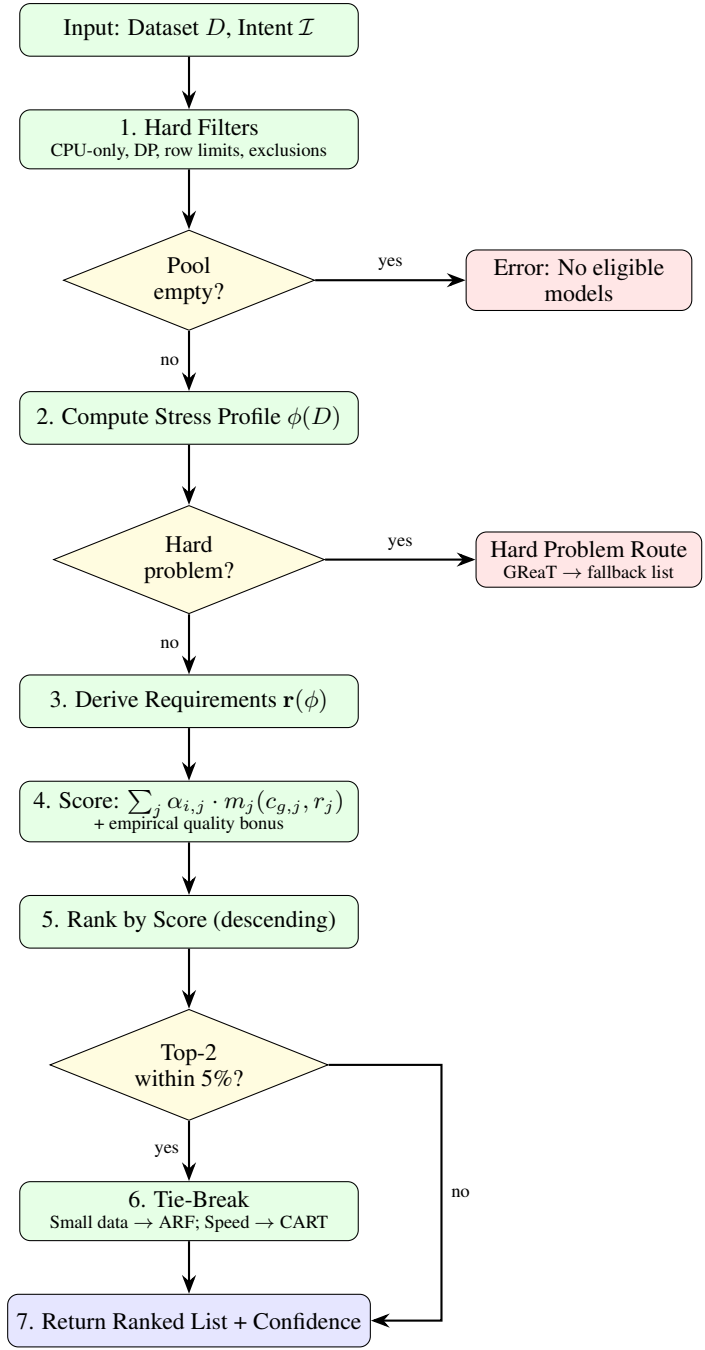

Figure~\ref{fig:rs-pipeline} illustrates the rule-based scoring pipeline as
a decision flowchart. The pipeline processes each dataset-intent pair through
seven sequential stages, with two early-exit paths (hard filters and hard
problem routing).

\newpage

\paragraph{Stage 1: Hard Filters.}
Models are excluded based on static constraints: \texttt{exclude} flag (baselines like Identity), \texttt{cpu\_only} compatibility (removes 6 GPU-required models), \texttt{strict\_dp}
certification (retains only PATECTGAN, AIM, DPCART with $\text{privacy\_dp} \geq 3$), and row-count limits (\texttt{min\_rows}, \texttt{max\_recommended\_rows}).

\paragraph{Stage 2: Hard-Problem Detection.}
A dataset is classified as a ``hard problem'' when \emph{all three} stress conditions co-occur: severe skew ($|\text{skewness}| > 2.0$), high cardinality ($> 500$ unique values), and Zipfian concentration (top-20\% categories $> 5\%$ of mass). Hard problems are routed directly to a designated model (GReaT, or TabDDPM for large datasets), bypassing the general scoring path.

\paragraph{Stages 3--4: Requirement Derivation and Scoring.}
Each stress factor maps to a required capability level via threshold-based quantization (Section~\ref{sec:stress-profiling}). The match function applies the score decay curve (Table~\ref{tab:decay-curve}), weighted by intent-conditioned scale factors when provided. An empirical quality bonus ($\text{avg\_quality\_score} \times 0.3$) from Spark benchmarks is added to
each model's total score.

\begin{table}[ht]
\centering
\caption{Score decay curve: match score as a function of capability gap.}
\label{tab:decay-curve}
\begin{tabular}{lcc}
\toprule
Condition & Match Score & Interpretation \\
\midrule
$c_{g,j} \geq r_j$ & 1.0 & Capability meets or exceeds requirement \\
$c_{g,j} = r_j - 1$ & 0.7 & Near miss (one level below) \\
$c_{g,j} = r_j - 2$ & 0.4 & Moderate gap (two levels below) \\
$c_{g,j} < r_j - 2$ & 0.0 & Insufficient capability \\
\bottomrule
\end{tabular}
\end{table}

\paragraph{Stages 5--7: Ranking, Tie-Breaking, and Output.}
Models are ranked by descending total score. When the top two models score within 5\%, tie-breaking rules apply in priority order: (1)~small datasets ($< 1{,}000$ rows) prefer ARF; (2)~speed-optimized intents prefer CART; (3)~quality-optimized intents prefer diffusion models (GPU) or tree models (CPU). The final output includes the primary recommendation with a confidence score, up to $N$ alternatives, and a difficulty summary.

\subsection{API Implementation and Persistence Layer}
\label{sec:appendix-api}

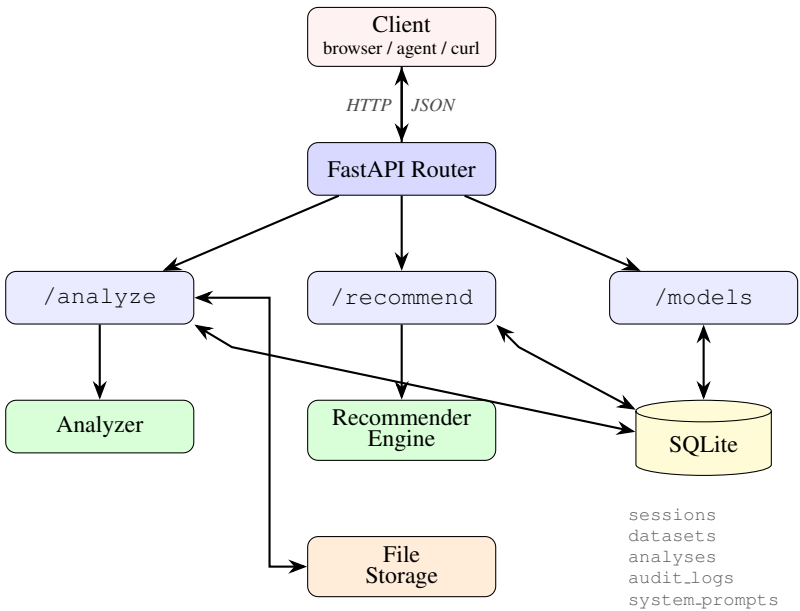
\begin{figure}[ht]
\centering
\begin{tikzpicture}[
    node distance=0.6cm and 1.0cm,
    box/.style={draw, rounded corners, minimum width=2.5cm, minimum height=0.7cm,
                align=center, font=\small},
    db/.style={draw, cylinder, shape border rotate=90, minimum width=1.8cm,
               minimum height=1.0cm, aspect=0.25, align=center, font=\small},
    arr/.style={-{Stealth[length=2.5mm]}, thick},
    darr/.style={{Stealth[length=2.5mm]}-{Stealth[length=2.5mm]}, thick},
    lbl/.style={font=\scriptsize\itshape, text=gray!60!black},
]

\node[box, fill=pink!20] (client) {Client\\[-2pt]{\scriptsize browser / agent / curl}};

\node[box, fill=blue!15, below=1.0cm of client] (api) {FastAPI Router};

\node[box, fill=blue!8, below left=1.0cm and 1.5cm of api] (ep1) {\texttt{/analyze}};
\node[box, fill=blue!8, below=1.0cm of api] (ep2) {\texttt{/recommend}};
\node[box, fill=blue!8, below right=1.0cm and 1.5cm of api] (ep3) {\texttt{/models}};

\node[box, fill=green!15, below=1.0cm of ep1] (analyzer) {Analyzer};
\node[box, fill=green!15, below=1.0cm of ep2] (engine) {Recommender\\[-2pt]Engine};

\node[db, fill=yellow!20, below=1.0cm of ep3] (sqlite) {SQLite};

\node[box, fill=orange!15, below=1.0cm of engine] (storage) {File\\[-2pt]Storage};

\node[font=\scriptsize, below=0.3cm of sqlite, align=left, text=gray!70!black] (tables) {
    \texttt{sessions}\\
    \texttt{datasets}\\
    \texttt{analyses}\\
    \texttt{audit\_logs}\\
    \texttt{system\_prompts}
};

\draw[arr] (client) -- node[left, lbl] {HTTP} (api);
\draw[arr] (api) -- (ep1);
\draw[arr] (api) -- (ep2);
\draw[arr] (api) -- (ep3);
\draw[arr] (ep1) -- (analyzer);
\draw[arr] (ep2) -- (engine);
\draw[darr] (ep1.south east) -- ++(0.5, -0.3) -- (sqlite);
\draw[darr] (ep2.south east) -- ++(0.3, -0.3) -- (sqlite);
\draw[darr] (ep3) -- (sqlite);
\draw[darr] (ep1.east) -- ++(1.0,0) |- (storage.west);
\draw[arr] (api) -- node[right, lbl] {JSON} (client);

\end{tikzpicture}
\caption{API architecture with SQLite persistence. Double arrows indicate read/write paths. Uploaded files are stored on disk; profiles and recommendations are cached in SQLite for session-based retrieval.}
\label{fig:api-architecture}
\end{figure}

\textsc{SYNTHONY} exposes a REST API via FastAPI with a SQLite persistence layer that
caches dataset profiles, tracks user sessions, and versions system prompts
(Figure~\ref{fig:api-architecture}).

\paragraph{Database Schema.}
The persistence layer uses five SQLAlchemy ORM tables (Table~\ref{tab:db-schema}). Sessions have a 30-day retention policy with automatic cleanup. Analyses cache serialized \texttt{DatasetProfile} and \texttt{ColumnAnalysisResult} as JSON, enabling the \texttt{/recommend} endpoint to retrieve cached profiles by \texttt{analysis\_id} without reprocessing.

\begin{table}[ht]
\centering
\caption{SQLite database schema. PK = primary key, FK = foreign key.}
\label{tab:db-schema}
\resizebox{\linewidth}{!}{%
\begin{tabular}{llll}
\toprule
Table & Key Columns & Relationships & Purpose \\
\midrule
\texttt{sessions} & \texttt{session\_id} (PK), \texttt{ip\_address}, \texttt{expires\_at} & $\to$ datasets (1:N) & Track client sessions (30-day TTL) \\
\texttt{datasets} & \texttt{dataset\_id} (PK), \texttt{session\_id} (FK), \texttt{file\_path} & $\to$ analyses (1:N) & Uploaded file metadata \\
\texttt{analyses} & \texttt{analysis\_id} (PK), \texttt{dataset\_id} (FK), \texttt{profile\_json} & $\to$ system\_prompts & Cached profiles + recommendations \\
\texttt{system\_prompts} & \texttt{prompt\_id} (PK), \texttt{version}, \texttt{is\_active} & --- & Versioned LLM system prompts \\
\texttt{audit\_logs} & \texttt{log\_id} (PK), \texttt{session\_id}, \texttt{action} & --- & Request audit trail \\
\bottomrule
\end{tabular}%
}
\end{table}

\paragraph{Request Flow.}
A typical \texttt{POST /analyze-and-recommend} request follows this path:
\begin{enumerate}
    \item \textbf{Session creation:} Generate UUID; store client IP and user-agent with 30-day expiry.
    \item \textbf{File persistence:} Save uploaded CSV/Parquet to \texttt{data/uploads/\{session\_id\}/}, enforcing per-file (100\,MB), per-session (500\,MB), and total (10\,GB) storage quotas.
    \item \textbf{Profiling:} Run \texttt{StochasticDataAnalyzer.analyze(df)} and \texttt{ColumnAnalyzer.analyze(df)} to produce the stress profile and column analysis.
    \item \textbf{Cache:} Serialize results as JSON and insert into \texttt{analyses} table. Subsequent \texttt{/recommend} calls can reference the cached \texttt{analysis\_id} to skip reprofiling.
    \item \textbf{Recommendation:} Pass the profile to \texttt{ModelRecommendationEngine.recommend()} with the specified method and constraints.
    \item \textbf{Audit:} Log the action, endpoint, IP, and success/failure to \texttt{audit\_logs}.
\end{enumerate}
\paragraph{Endpoint Summary.}
Table~\ref{tab:api-endpoints} lists the primary API endpoints.

\begin{table}[ht]
\centering
\caption{REST API endpoint summary.}
\label{tab:api-endpoints}
\begin{tabular}{llp{6cm}}
\toprule
Method & Endpoint & Description \\
\midrule
\texttt{POST} & \texttt{/analyze} & Upload CSV/Parquet $\to$ stress profile + column analysis (cached) \\
\texttt{POST} & \texttt{/recommend} & Profile $\to$ ranked recommendation (rule/LLM/hybrid) \\
\texttt{POST} & \texttt{/analyze-and-recommend} & One-shot: upload $\to$ profile $\to$ recommendation \\
\texttt{GET} & \texttt{/models} & List models with optional filters (type, CPU, DP) \\
\texttt{GET} & \texttt{/models/\{name\}} & Detailed model capabilities and constraints \\
\texttt{GET} & \texttt{/health} & Component status (analyzer, recommender, LLM availability) \\
\bottomrule
\end{tabular}
\end{table}

\subsection{MCP Server: Agentic Integration}
\label{sec:appendix-mcp}

\begin{figure}[ht]
\centering
\begin{tikzpicture}[
    node distance=0.6cm and 0.8cm,
    box/.style={draw, rounded corners, minimum width=2.4cm, minimum height=0.7cm,
                align=center, font=\small},
    tool/.style={draw, rounded corners, minimum width=2.2cm, minimum height=0.6cm,
                 align=center, font=\scriptsize, fill=green!10},
    res/.style={draw, rounded corners, minimum width=2.2cm, minimum height=0.6cm,
                align=center, font=\scriptsize, fill=yellow!10},
    prompt/.style={draw, rounded corners, minimum width=2.2cm, minimum height=0.6cm,
                   align=center, font=\scriptsize, fill=blue!10},
    arr/.style={-{Stealth[length=2.5mm]}, thick},
    darr/.style={{Stealth[length=2.5mm]}-{Stealth[length=2.5mm]}, thick},
    lbl/.style={font=\scriptsize\itshape},
]

\node[box, fill=pink!20] (agent) {AI Agent\\[-2pt]{\scriptsize Claude / LangChain}};

\node[box, fill=blue!15, below=1.2cm of agent] (mcp) {MCP Server\\[-2pt]{\scriptsize JSON-RPC 2.0 / stdio}};

\node[font=\small\bfseries, below left=1.2cm and 3cm of mcp] (tlbl) {Tools (13)};
\node[tool, below=0.2cm of tlbl] (t1) {\texttt{analyze\_stress\_profile}};
\node[tool, below=0.15cm of t1] (t2) {\texttt{rank\_models\_hybrid}};
\node[tool, below=0.15cm of t2] (t3) {\texttt{rank\_models\_rule}};
\node[tool, below=0.15cm of t3] (t4) {\texttt{list\_datasets}};
\node[tool, below=0.15cm of t4] (t5) {\texttt{benchmark\_compare}};
\node[tool, below=0.15cm of t5] (t6) {\texttt{...}};

\node[font=\small\bfseries, below=1.2cm of mcp] (rlbl) {Resources (5)};
\node[res, below=0.2cm of rlbl] (r1) {\texttt{models://registry}};
\node[res, below=0.15cm of r1] (r2) {\texttt{models://model/\{name\}}};
\node[res, below=0.15cm of r2] (r3) {\texttt{datasets://profiles/\{id\}}};
\node[res, below=0.15cm of r3] (r4) {\texttt{benchmarks://thresholds}};
\node[res, below=0.15cm of r4] (r5) {\texttt{guidelines://system-prompt}};

\node[font=\small\bfseries, below right=1.2cm and 3cm of mcp] (plbl) {Prompts (4)};
\node[prompt, below=0.2cm of plbl] (p1) {\texttt{analyze-and-recommend}};
\node[prompt, below=0.15cm of p1] (p2) {\texttt{explain-hard-problem}};
\node[prompt, below=0.15cm of p2] (p3) {\texttt{validate-recommendation}};
\node[prompt, below=0.15cm of p3] (p4) {\texttt{update-knowledge-base}};

\node[box, fill=green!12, below=0.8cm of t6] (core) {Synthony Core\\[-2pt]{\scriptsize Analyzer + Engine}};

\draw[darr] (agent) -- node[right, lbl] {JSON-RPC} (mcp);
\draw[arr] (mcp) -- (tlbl);
\draw[arr] (mcp) -- (rlbl);
\draw[arr] (mcp) -- (plbl);
\draw[arr] (t6) -- (core);

\end{tikzpicture}
\caption{MCP server architecture. The server exposes SYNTHONY's capabilities
  through three MCP primitives: \emph{Tools} (executable functions),
  \emph{Resources} (read-only data), and \emph{Prompts} (guided workflows).
  Communication uses JSON-RPC~2.0 over stdio.}
\label{fig:mcp-architecture}
\end{figure}
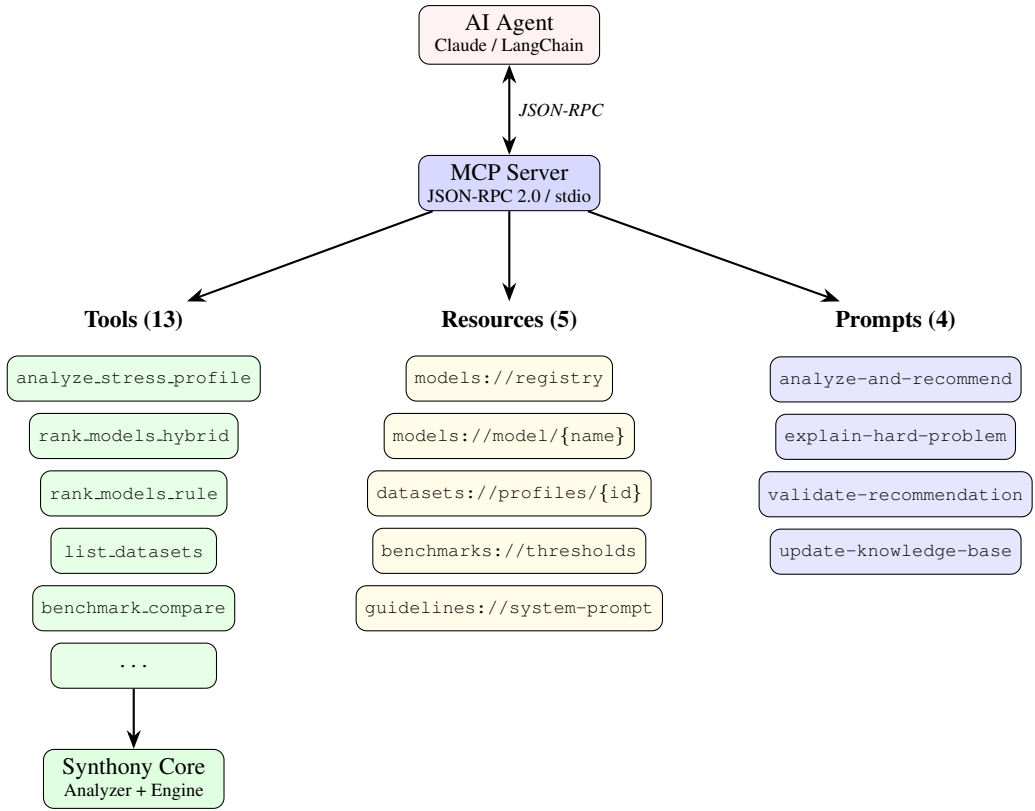

\textsc{SYNTHONY} implements a Model Context Protocol (MCP) server that enables AI
agents (e.g., Claude Code, LangChain) to invoke profiling and recommendation
tools programmatically via JSON-RPC~2.0 over stdio transport
(Figure~\ref{fig:mcp-architecture}).

\paragraph{MCP Primitives.}
The MCP specification defines three primitive types that map naturally to
\textsc{SYNTHONY}'s capabilities:

\begin{itemize}
    \item \textbf{Tools} are executable functions that the agent can invoke
      with structured arguments. \textsc{SYNTHONY} registers 13 tools spanning data
      loading (\texttt{list\_datasets}, \texttt{load\_dataset}), profiling
      (\texttt{analyze\_stress\_profile}), recommendation
      (\texttt{rank\_models\_hybrid}, \texttt{rank\_models\_rule},
      \texttt{rank\_models\_llm}), model inspection
      (\texttt{get\_model\_info}, \texttt{list\_models},
      \texttt{check\_model\_constraints}), explanation
      (\texttt{explain\_recommendation\_reasoning},
      \texttt{get\_tie\_breaker\_logic}), and benchmarking
      (\texttt{benchmark\_compare}, \texttt{generate\_benchmark\_dataset}).

    \item \textbf{Resources} are read-only data endpoints identified by URIs. The server exposes the full model registry (\texttt{models://registry}), individual model details (\texttt{models://model/\{name\}}), cached dataset profiles (\texttt{datasets://profiles/\{id\}}), stress detection thresholds (\texttt{benchmarks://thresholds}), and the active LLM system prompt (\texttt{guidelines://system-prompt}).

    \item \textbf{Prompts} are guided multi-step workflows. Four prompts
      encode common task sequences: \texttt{analyze-and-recommend} (full
      pipeline from data path to explanation),
      \texttt{explain-hard-problem} (deep dive into stress factors),
      \texttt{validate-recommendation} (offline benchmark validation), and
      \texttt{update-knowledge-base} (empirical feedback loop to refine
      capability scores).
\end{itemize}

\paragraph{Agentic Workflow.}
An AI agent interacting with \textsc{SYNTHONY} through MCP follows a typical
three-step workflow:

\begin{enumerate}
    \item \textbf{Discovery:} The agent calls \texttt{tools/list} and
      \texttt{resources/list} to discover available capabilities.
    \item \textbf{Profiling:} The agent invokes
      \texttt{analyze\_stress\_profile} with a dataset path, receiving the
      stress profile and column analysis as structured JSON.
    \item \textbf{Selection:} The agent passes the profile to
      \texttt{rank\_models\_hybrid}, optionally specifying constraints
      (\texttt{cpu\_only}, \texttt{strict\_dp}) and the number of
      alternatives. The response includes the ranked recommendation,
      confidence scores, and decision reasoning.
\end{enumerate}

This workflow can be orchestrated by a prompt template
(\texttt{analyze-and-recommend}) or composed ad hoc by the agent.

\paragraph{Example MCP Commands.}
Listing~\ref{lst:mcp-examples} shows representative JSON-RPC commands for the
three most common operations.

\begin{lstlisting}[
    caption={Example MCP JSON-RPC commands.},
    label={lst:mcp-examples},
    language={},
    basicstyle=\ttfamily\scriptsize,
    breaklines=true,
    frame=single,
    xleftmargin=2pt,
    xrightmargin=2pt,
]
// 1. List available datasets
{"jsonrpc":"2.0", "method":"tools/call", "id":1,
 "params":{"name":"list_datasets",
   "arguments":{"format_filter":"csv"}}}

// 2. Profile a dataset
{"jsonrpc":"2.0", "method":"tools/call", "id":2,
 "params":{"name":"analyze_stress_profile",
   "arguments":{"dataset_name":"Bean"}}}

// 3. Get ranked recommendations
{"jsonrpc":"2.0", "method":"tools/call", "id":3,
 "params":{"name":"rank_models_hybrid",
   "arguments":{
     "dataset_profile": <output from step 2>,
     "method":"hybrid", "top_n":3}}}

// 4. Read model registry (resource)
{"jsonrpc":"2.0", "method":"resources/read", "id":4,
 "params":{"uri":"models://registry"}}

// 5. Invoke guided workflow (prompt)
{"jsonrpc":"2.0", "method":"prompts/get", "id":5,
 "params":{"name":"analyze-and-recommend",
   "arguments":{"data_path":"/data/insurance.csv"}}}
\end{lstlisting}

\paragraph{Self-Correcting Feedback Loop.}
The \texttt{update-knowledge-base} prompt enables a feedback cycle: after
validating a recommendation against benchmark results, the agent can propose
updates to the capability registry or system prompt. This loop is currently
human-supervised (the agent proposes changes; a developer reviews and
applies them), but the architecture supports fully autonomous calibration
via the offline Bayesian optimization described in
Appendix~\ref{sec:appendix-optimization}.
\section{Implementation Details}
\label{sec:implementation}
We standardize the training and evaluation interface across all candidate synthesizers via a unified API exposing \texttt{fit()}, \texttt{sample()}, and \texttt{get\_hyperparameter\_space()} methods, enabling controlled comparisons under identical preprocessing and sampling conditions. An evaluation harness computes a metric vector $\mathbf{m} \in \mathbb{R}^d$ spanning statistical fidelity (e.g., column shape score), privacy risk (e.g., proportion of synthetic records closer to real data than holdout records), and utility (downstream ML task performance, measured by Test ROC AUC for classification and Test Adjusted R$^2$ for regression). An LLM can optionally translate natural-language intent into a scalarized objective, but the core selection is performed by the stress-aware scoring function described in Section~\ref{sec:method}.

\section{Benchmark Datasets}
\label{sec:dataset}
We provide the URL for the sources of each downstream benchmark set considered in the paper. 

\begin{enumerate}
    \item \textbf{Abalone} (OpenML) : \url{https://www.openml.org/search?type=data\&sort=runs\&id=183\&status=active} (multi-class)
    \item \textbf{Bean} (UCI) : \url{https://archive.ics.uci.edu/dataset/602/dry+bean+dataset} (Multi class)
    \item \textbf{faults} (UCI) : \url{https://archive.ics.uci.edu/dataset/198/steel+plates+faults} (multi-class)
    \item \textbf{IndianLiverPatient} (Kaggle) : \url{https://www.kaggle.com/datasets/uciml/indian-liver-patient-records} (binary)
    \item \textbf{insurance} (Kaggle) : \url{https://www.kaggle.com/datasets/mirichoi0218/insurance} (Regression)
    \item \textbf{Obesity} (Kaggle) : \url{https://www.kaggle.com/datasets/tathagatbanerjee/obesity-dataset-uci-ml} (multi-class)
    \item \textbf{wilt} (OpenML) : \url{https://www.openml.org/search?type=data\&sort=runs\&id=40983\&status=active} (binary)
\end{enumerate}

\section{Evaluation Metrics}

The three intent-specific metrics used to construct oracle rankings in Section~\ref{sec:experiments} are as follows.

\subsection{Fidelity: Column Shape Score}

We measure fidelity via the \textbf{Column Shape Score}, which evaluates similarity between real and synthetic marginal distributions on a per-column basis. For numerical columns, we use the Kolmogorov--Smirnov (KS) test statistic subtracted from one; for categorical columns, we use one minus the Total Variation Distance (TVD). The final score is the average across all columns, with higher values indicating better fidelity.

\subsection{Utility: Downstream ML Performance}

We evaluate machine learning utility by training classifiers or regressors on synthetic data and evaluating on held-out real test data. For classification datasets we report \textbf{test ROC AUC}; for regression datasets we report \textbf{test adjusted R$^2$}. Higher values indicate better utility in both cases.

\subsection{Privacy: Proportion Closer to Real}

We evaluate privacy risk via \textbf{Proportion Closer to Real}~\citep{platzer2021holdout}, which measures the fraction of synthetic records whose nearest real neighbor is in the training set rather than the holdout set. Lower values indicate better privacy (less memorization risk). For oracle ranking construction, we invert this metric so that higher rank corresponds to better privacy.

\section{Code and Data Availability}
\label{sec:appendix-code}

The complete source code for \textsc{Synthony}, including the profiling
pipeline, recommendation engine, capability registry, optimization scripts,
REST API, and MCP server, is publicly available at:
\begin{center}
\url{https://github.com/UCLA-Trustworthy-AI-Lab/Synthony}
\end{center}
And, the source code for the Synthesizer has refers to as:
\begin{center}
\url{https://github.com/UCLA-Trustworthy-AI-Lab/table-synthesizers}
\end{center}

\noindent The repository is organized as follows:

\begin{table}[H]
\centering
\caption{Repository structure and correspondence to paper sections.}
\label{tab:repo-structure}
\resizebox{\linewidth}{!}{%
\begin{tabular}{lll}
\toprule
Directory / File & Contents & Paper Reference \\
\midrule
\texttt{src/synthony/core/}
    & Data loaders, \texttt{StochasticDataAnalyzer}, schemas
    & Section~\ref{sec:stress-profiling} \\
\texttt{src/synthony/detectors/}
    & Skewness, cardinality, Zipfian, correlation detectors
    & Section~\ref{sec:stress-profiling} \\
\texttt{src/synthony/recommender/}
    & \texttt{ModelRecommendationEngine}, scoring logic
    & Section~\ref{sec:candidate-selection} \\
\texttt{config/model\_capabilities.json}
    & Capability registry (v7.0.0, 15 models $\times$ 6 dims)
    & Section~\ref{sec:candidate-selection} \\
\texttt{scripts/optimize\_scaling.py}
    & Bayesian optimization of scale factors (Optuna/TPE)
    & Section~\ref{sec:appendix-optimization} \\
\texttt{src/synthony/api/}
    & FastAPI REST server with SQLite persistence
    & Appendix~\ref{sec:appendix-api} \\
\texttt{mcp\_server/}
    & MCP server (JSON-RPC 2.0 / stdio)
    & Appendix~\ref{sec:appendix-mcp} \\
\texttt{src/synthony/benchmark/}
    & Data quality metrics (KL/JS divergence, fidelity, privacy)
    & Section~\ref{sec:experiments} \\
\texttt{tests/}
    & Unit and integration test suites
    & --- \\
\bottomrule
\end{tabular}%
}
\end{table}

\paragraph{Reproducing Results.}
To reproduce the experiments reported in this paper:

\begin{lstlisting}[
    language={},
    basicstyle=\ttfamily\scriptsize,
    breaklines=true,
    frame=single,
    xleftmargin=2pt,
    xrightmargin=2pt,
]
# 1. Install Synthony with all dependencies
pip install -e ".[all]"

# 2. Profile a dataset
synthony-profile dataset/input_data/Bean.csv --verbose

# 3. Run rule-based recommendation
synthony-recommender -i dataset/input_data/Bean.csv --method rulebased

# 4. Run the Bayesian optimization (scale factors)
python scripts/optimize_scaling.py

# 5. Start the API server
uvicorn synthony.api.server:app --reload

# 6. Start the MCP server (for AI agent integration)
python -m mcp_server.server --verbose
\end{lstlisting}

\noindent Full installation instructions, environment setup, and dataset
preparation steps are documented in the repository's \texttt{README.md}.
The benchmark datasets (Abalone, Bean, IndianLiverPatient, Obesity, faults,
insurance, wilt) are sourced from OpenML~\citep{vanschoren2013openml}.

\end{document}